\documentclass[10pt]{article}
\usepackage[preprint]{tmlr}
\usepackage{amsmath,amssymb,booktabs,graphicx,tabularx,longtable}
\usepackage{hyperref,url}
\hypersetup{hidelinks,pdftitle={What Changes When Fact-Verification Scores Improve? Evidence and Answer Accounting Across Trained Verifiers and LLMs},pdfauthor={Han Chen; Yingrui Li},pdfsubject={Research preprint}}
\title{What Changes When Fact-Verification Scores Improve?\\Evidence and Answer Accounting Across Trained Verifiers and LLMs\thanks{The authors originated the main research question and the two study designs. ChatGPT, Claude, and Codex assisted with implementation, execution, automated checks, interpretation, and manuscript drafting and revision, and suggested some post-hoc analyses.}}
\author{Han Chen\thanks{Both authors contributed equally and share first authorship.}\AND Yingrui Li\footnotemark[2]}
\newcommand{\invalid}{\texttt{INVALID}}
\newcommand{\qwen}{Qwen3-8B}
\newcommand{\llama}{Llama-3.1-8B-Instruct}
\newcommand{\LegacyDD}{32.41}

\newcommand{\LegacyUU}{42.02}
\newcommand{\LegacyGain}{9.61}
\newcommand{\LegacyEzero}{7.92}
\newcommand{\LegacyAone}{1.69}
\newcommand{\LegacyAzero}{0.53}
\newcommand{\LegacyEone}{9.08}
\newcommand{\MayMacroD}{49.34}
\newcommand{\MayMacroH}{54.16}
\newcommand{\MayStrictD}{35.16}
\newcommand{\MayStrictH}{34.78}
\newcommand{\MayNeiH}{17.67}
\newcommand{\LegacyGainLo}{8.77}
\newcommand{\LegacyGainHi}{10.43}
\newcommand{\LegacySubsetGain}{9.42}
\newcommand{\LegacySubsetLo}{8.62}
\newcommand{\LegacySubsetHi}{10.25}

\begin{document}
\maketitle

\begin{abstract}
A joint fact-verification score assesses answers and submitted evidence together. When the score improves, how much of the gain remains if the answers are held fixed? On \mbox{FEVEROUS}, strict score is the percentage of claims with a correct answer and a complete annotated evidence group in the submitted evidence. Across four trained DeBERTa checkpoints and 7,890 claims, replacing DCUF evidence with UnifEE evidence raises strict score by 9.61 percentage points, compared with 1.96 percentage points in answer accuracy. The paired 95\% interval for the strict-score gain is [8.77, 10.43], conditional on these checkpoints. Replacing only the evidence passed to the scorer accounts for 7.92 or 9.08 percentage points when we retain the answers generated from DCUF or UnifEE evidence, respectively. To examine how this evidence gain depends on evaluation choices, we generate 470,400 responses from two 8B LLMs on FEVER, FEVEROUS and SciFact under two answer formats and two context budgets. Increasing context from 256 to 2,048 tokens raises the fixed-answer evidence gain on FEVEROUS by 3.84 and 3.10 percentage points for Qwen and Llama, respectively. The effects fall short of the prespecified cross-dataset criterion, while some intervals extend beyond the two-point small-effect bound. Post-hoc analyses quantify changes in answers and submitted evidence, and show when aggregate accuracy and evidence-coverage rates miss the claim-level pattern. The four answer--evidence score combinations reveal changes that endpoint and aggregate metrics leave unresolved.
\end{abstract}

\section{Introduction}
A fact-verification system returns an answer and supporting evidence. Joint benchmarks award credit when both satisfy the scoring requirements \citep{thorne2018fever,aly2021feverous}. A higher score can reflect better answers, better evidence, or a change in which claims have both. The total gain leaves these contributions unresolved.

For example, \citet[][Table~2 and Section~3.4]{hu2023unifee} report UnifEE development-set gains over DCUF of 9.09 percentage points in FEVEROUS score and 0.76 in answer accuracy, alongside evidence-retrieval metrics. We examine how much joint gain survives at fixed answers by recombining saved answers and evidence submissions.

In Study~1, switching from DCUF to UnifEE evidence with each trained checkpoint held fixed raises mean strict score by 9.61 percentage points and answer accuracy by 1.96. Keeping the answers generated from DCUF evidence fixed while replacing only the scorer's evidence retains 7.92 percentage points; updating the answers adds 1.69. Reversing the steps assigns 9.08 percentage points to evidence and 0.53 to answers.

The next question is whether this evidence gain depends on how much evidence a model sees or how it returns an answer. Study~2 tests this with fixed-weight LLMs, two formats and two context budgets, using BM25 and BGE evidence orderings on three benchmarks. FEVEROUS shows context dependence; the other datasets fall short of the prespecified replication criterion.

\paragraph{Contributions.}
Our empirical contribution is to quantify the evidence component of a source gain, test its sensitivity to LLM interfaces, and explain when aggregate rates recover it. The $2\times2$ comparison reveals a concrete contrast: marginal accuracy and evidence coverage closely approximate one trained-verifier allocation but miss most of the FEVER/Qwen context effect. The missing information is which claims combine correct answers with eligible evidence.

A separate retrospective example illustrates the related trade-off between class metrics: a Not Enough Information (NEI) override improves macro-F1 but lowers answer accuracy and strict score. We retain its reconstruction and unresolved threshold history in Appendix~\ref{app:legacy}.

The $2\times2$ comparison requires saving answers and submitted evidence under both source conditions. Two additional scoring passes provide the off-diagonal scores and complete the comparison.

\section{Related Work}
\paragraph{Answer and evidence evaluation.}
FEVER combines claim labels with Wikipedia evidence \citep{thorne2018fever}; FEVEROUS adds structured sources and requires evidence for NEI claims \citep{aly2021feverous}. SciFact evaluates scientific claims through abstract labels and rationale sentences \citep{wadden2020scifact}. DCUF develops text--table fusion, while UnifEE improves evidence extraction and reports both answer accuracy and joint scores \citep{hu2022dcuf,hu2023unifee}. We use their evidence outputs with our own trained checkpoints. Our question concerns the score changes under this intervention, rather than reproducing their original end-to-end systems. In Study~2, BM25 \citep{robertson2009bm25} and a BGE reranker \citep{xiao2023cpack} reorder the same candidate pools.

\paragraph{Interfaces and evidence use.}
Prompt formatting and option identifiers can change model outputs \citep{sclar2024format,zheng2024selectors}; strict answer matching can combine semantic errors with output-format failures \citep{hua2025artifact}. Long-context performance also depends on information placement \citep{liu2024lost}, including in retrieval-augmented fact checking \citep{bernardelle2026context}. We study how these evaluation choices affect the gain from replacing scored evidence at fixed answers. Attribution and citation work similarly distinguishes response correctness from evidential support \citep{buchmann2024attribute,gao2023citations}. Claim-only artifacts in FEVER show why answer correctness alone is insufficient to establish evidence use \citep{schuster2019debiasing}. Our LLM experiments use Qwen3 and Llama~3 models \citep{qwen2025qwen3,grattafiori2024llama}.

\paragraph{Decomposition and uncertainty.}
Order-dependent decompositions and averaging over orders are established methods \citep{shorrocks2013decomposition}. Averaging the two evidence components here gives the two-player Shapley allocation. We contribute empirical comparisons of these components, not a new attribution identity. Paired inference follows each metric's dependence structure \citep{dror2018significance}. Study~2 resamples claim families and decoding seeds, resembling crossed-factor resampling \citep{owen2007bootstrap}; it estimates uncertainty for fixed models. We distinguish an interval that includes zero from evidence that an effect is small \citep{card2020power}.

\section{Decomposing Answer and Evidence Changes}
\label{sec:accounting}
We call the predicted class the \emph{answer}. An \emph{official score} uses the benchmark's own scoring implementation and submission limits. \emph{Evidence eligibility} means that a submitted evidence set satisfies its evidence requirement for the gold answer. This concerns annotated evidence identifiers; semantic relevance and the evidence visible to the model are separate properties.

We score each set of saved answers with evidence from both sources. Let $S_{ab}$ denote the resulting corpus score on a 0--100 scale: $a$ identifies the source used to generate answers, and $b$ identifies the submitted evidence. Changes are in percentage points for strict score and answer accuracy, and score points for F1; we use ``points'' as shorthand for both below. We build a separate table for each model, input condition and replicate. Source 0 is DCUF in Study~1 and BM25 in Study~2; source 1 is UnifEE and BGE, respectively. We generate answers separately under both sources. The diagonal entries are the observed endpoint scores, while the off-diagonal entries reuse those answers with the other evidence submission.

\begin{table}[htbp]
\caption{The $2\times2$ score table. Each row holds answers fixed; each column holds submitted evidence fixed.}
\label{tab:schematic}
\begin{center}
\begin{tabular}{lcc}
\toprule
Saved answers & Source 0 scoring evidence & Source 1 scoring evidence \\
\midrule
Generated with source 0 & $S_{00}$: original endpoint & $S_{01}$: evidence replaced \\
Generated with source 1 & $S_{10}$: answers replaced & $S_{11}$: new endpoint \\
\bottomrule
\end{tabular}
\end{center}
\end{table}

The endpoint gain $G$ has two exact decompositions:
\begin{align}
G &= S_{11}-S_{00} = \underbrace{(S_{01}-S_{00})}_{E_0}
                         +\underbrace{(S_{11}-S_{01})}_{A_1},\label{eq:first}\\
  &= \underbrace{(S_{10}-S_{00})}_{A_0}
                         +\underbrace{(S_{11}-S_{10})}_{E_1}.\label{eq:second}
\end{align}
$E_0$ is the \emph{evidence gain} with source-0 answers held fixed, and $E_1$ uses source-1 answers. The first order changes evidence then answers; the second changes answers then evidence. Their difference is the interaction
\begin{equation}
I = S_{11}-S_{10}-S_{01}+S_{00}=E_1-E_0.
\end{equation}
Averaging the orders assigns $(E_0+E_1)/2$ to evidence. These equalities hold for nonlinear corpus metrics as well as accuracy. They allocate score changes, not causal effects on reasoning. We use additive score points because component-to-total ratios are unstable when the total is small. The answers are fixed within each evidence comparison; another input condition may produce different answers.

\paragraph{When marginal rates suffice.}
For strict accuracy, the score of a claim is the product of answer correctness and evidence eligibility. The off-diagonal scores follow exactly from these two indicators. At input condition $c$, let $a_i(c)$ denote source-0 answer correctness, averaged over the reported replicates, and let $d_i(c)=q_{1i}(c)-q_{0i}(c)$ be the difference between the two evidence-eligibility indicators. Then
\begin{equation}
E_0(c)=100\,\overline{a(c)d(c)}
      =100\left[\overline{a(c)}\,\overline{d(c)}+\operatorname{Cov}_n(a(c),d(c))\right],
\label{eq:covariance}
\end{equation}
where the covariance uses denominator $n$. Multiplying marginal answer accuracy by the eligibility difference omits this covariance: whether eligibility changes on the same claims that the model answers correctly. We apply this diagnostic in both studies. SciFact uses corpus F1, so we score its four combinations directly rather than apply the strict-accuracy factorization.

\section{Study 1: Evidence Replacement with Trained Verifiers}
\label{sec:legacy}
\subsection{Design}
We evaluate four trained DeBERTa-large-MNLI checkpoints \citep{he2021deberta,williams2018mnli} on DCUF and UnifEE evidence for the same 7,890 FEVEROUS development claims. Each checkpoint generates answers from both sources. We hold its weights fixed, then score both answer vectors with both evidence submissions. For the FEVEROUS strict score, answer correctness and evidence eligibility factorize, so the off-diagonal values are exact recombinations of saved indicators.

All four checkpoints start from the same model revision and are fine-tuned on DCUF-derived inputs. Each sees 16,000 distinct training examples, updates all 406.2 million parameters, and is evaluated at the fixed final update of 1,000. Inputs are limited to 512 serialized tokens: a 64-token claim block and up to eight resolved evidence units, each capped at 96 tokens, selected from the first 30 candidate IDs. We include optional context when available, omit unresolved units, and use a sentinel for empty inputs. Scoring preserves provider order with at most five nonstructured and 25 structured items. Appendix~\ref{app:legacy} gives the full training and input specification.

The comparison controls verifier weights while changing the evidence source. It also changes content, ordering and the distribution of inputs relative to training. We therefore interpret the score split for these DCUF-trained checkpoints. The development population had been inspected previously, making this a retrospective comparison under a fixed execution rule.

\subsection{A Large Joint Gain with a Smaller Accuracy Gain}
\begin{table}[htbp]
\caption{Study 1 mean FEVEROUS strict scores over four trained checkpoints. Each row uses answers generated from the named evidence source and scores them with either evidence submission.}
\label{tab:legacy-cells}
\begin{center}
\begin{tabular}{lrr}
\toprule
Input evidence & DCUF scoring evidence & UnifEE scoring evidence \\
\midrule DCUF & 32.41 & 40.33 \\
UnifEE & 32.94 & 42.02 \\
\bottomrule
\end{tabular}
\end{center}
\end{table}

Strict score rises from \LegacyDD{} to \LegacyUU{}, a mean gain of \LegacyGain{} points with paired 95\% interval [\LegacyGainLo{}, \LegacyGainHi{}]. Answer accuracy increases from 68.09 to 70.05 (1.96 points), and macro-F1 increases by 1.44 points. The four cells in Table~\ref{tab:legacy-cells} identify the part of the larger joint gain that survives at fixed answers.

Changing evidence first contributes $E_0=\LegacyEzero{}$ points; changing answers next contributes $A_1=\LegacyAone{}$. Reversing the order assigns $A_0=\LegacyAzero{}$ points to answers and $E_1=\LegacyEone{}$ to evidence. Table~\ref{tab:study1-intervals} gives the component intervals. The 1.15-point difference between evidence gains reflects the interaction between which answers are correct and which submissions are eligible.
\begin{table}[htbp]
\caption{Study 1 score decomposition in points. All six quantities are contrasts of the same four score cells. Pointwise 95\% intervals use 10,000 paired claim-bootstrap draws conditional on the four trained checkpoints. The five component/interaction intervals were added post hoc; the total-gain interval reproduces the original analysis.}
\label{tab:study1-intervals}
\begin{center}
\begin{tabular}{llrr}
\toprule
Quantity & Change & Estimate & 95\% interval \\
\midrule
$G$ & Total source gain & 9.61 & [8.77, 10.43] \\
$E_0$ & Evidence replaced first & 7.92 & [7.18, 8.66] \\
$A_1$ & Answers replaced second & 1.69 & [1.28, 2.09] \\
$A_0$ & Answers replaced first & 0.53 & [0.20, 0.87] \\
$E_1$ & Evidence replaced second & 9.08 & [8.32, 9.83] \\
$I=E_1-E_0$ & Difference between orders & 1.15 & [0.86, 1.45] \\
\bottomrule
\end{tabular}
\end{center}
\end{table}

We use 10,000 paired claim-bootstrap draws with the original RNG seed 20261099. Each draw shares its claim sample across both sources and all four checkpoints. The five component/interaction intervals are post-hoc, pointwise 95\% summaries conditional on those checkpoints; the total-gain interval reproduces the original analysis. They describe each estimate separately and carry no simultaneous-coverage guarantee. A claim is the sampling unit; the 31,560 checkpoint/claim rows reuse each claim four times.

Every checkpoint predicts only SUPPORTS or REFUTES. The 501 NEI-gold claims consequently contribute zero to all four strict scores. The entire gain comes from 3,908 SUPPORTS and 3,481 REFUTES claims, contributing 5.93 and 3.69 points respectively on the 7,890-claim denominator (Appendix~\ref{app:legacy}). The decomposition remains exact, but its size need not describe a verifier that handles all three classes well.

The primary population includes 239 claims with an unresolved input unit under at least one source. Restricting to the predeclared 7,651 both-source-resolved claims gives a gain of \LegacySubsetGain{} [\LegacySubsetLo{}, \LegacySubsetHi{}]. All four checkpoint-specific endpoint gains are positive. Appendix~\ref{app:legacy} reports these sensitivities and the original metric intervals.

\subsection{How Much Do Marginal Rates Explain?}
The evidence-eligibility rate increases from 43.22\% to 55.07\%. These aggregate rates closely agree with the development-set evidence recalls reported by \citet[][Table~2]{hu2023unifee}: 43.22\% and 55.08\%, respectively. Multiplying our 11.85-point eligibility difference by source-0 answer accuracy (68.09\%) predicts an evidence gain of 8.07 points. Equation~\ref{eq:covariance} gives the residual: a covariance contribution of -0.14 points, yielding the observed 7.92. With source-1 answers, the corresponding product is 8.30 and the covariance contribution is 0.78, yielding 9.08.

Thus marginal rates approximate one order well but are less accurate for the other. The covariance records the overlap between correct answers and changing evidence eligibility. In Study~2 we use the same diagnostic to test whether marginal changes explain the context effect.

\subsection{From Source Gains to Interface Sensitivity}
Changing only the evidence passed to the scorer preserves 7.92 or 9.08 of the trained-verifier comparison's 9.61-percentage-point gain. Does this evidence gain depend on how much evidence a model sees or how it returns an answer? We examine that question with fixed-weight LLMs, BM25 and BGE evidence orderings, and three benchmarks (Table~\ref{tab:study-protocols}).
\begin{table}[htbp]
\caption{The two study designs. Training seeds in Study~1 and decoding seeds in Study~2 support different uncertainty statements; effects are reported separately. }
\label{tab:study-protocols}
\begin{center}
\begin{tabularx}{\textwidth}{>{\raggedright\arraybackslash}p{0.16\textwidth}>{\raggedright\arraybackslash}X>{\raggedright\arraybackslash}X}
\toprule
Dimension & Study 1: trained verifiers & Study 2: LLMs \\
\midrule
Models & Four trained DeBERTa checkpoints; each fixed across sources & Qwen3-8B and Llama-3.1-8B-Instruct; fixed weights \\
Evidence comparison & DCUF vs. UnifEE outputs & BM25 vs. BGE within fixed candidate pools \\
Evaluation population & 7,890 FEVEROUS development claims & 2,000 FEVER, 2,000 FEVEROUS and 300 SciFact claims \\
Input / scoring limits & 512 serialized tokens; up to 8 resolved units; scoring limits 5/25 & 256/2,048 evidence tokens; scoring limits; three SciFact abstracts \\
Metric / uncertainty & Strict score and class metrics; paired claim bootstrap conditional on fixed checkpoints & Strict score / rationalized F1; claim-family and decoding-seed bootstrap \\
Selection & Fixed final update; previously inspected development population & Internally frozen after development; historical exposure limitations \\
Retained records & Paired labels, fingerprints, coverage and replayable intervals; weights excluded & Raw responses, token/evidence IDs, counts, matrices and recorded plans \\
\bottomrule
\end{tabularx}
\end{center}
\end{table}

The studies also have distinct but historically related populations. Study~2's FEVEROUS panel is disjoint by ID from Study~1's development claims, but comes from a previously used checkpoint-validation pool. An earlier LLM development study using DCUF/UnifEE failed its quality gates and stopped before primary evaluation. The completed BM25/BGE study follows its own later protocol. Appendix~\ref{app:development} records this history and the population joins.

\section{Study 2: Frozen LLM Evaluation}\label{sec:llm}
We use $E_0$, the evidence gain with answers generated from BM25 evidence held fixed, as the primary outcome. We specified the format and context contrasts and the decision rule before generating the primary responses; the later analyses examine the resulting patterns.

\subsection{Data and Sampling Units}
Table~\ref{tab:data} gives the evaluation populations. We select 2,000 FEVER development claims by a deterministic hash and 2,000 eligible FEVEROUS claims from a historical verifier-validation pool. The FEVEROUS selection excludes families linked to reserved or previously inspected development cases. SciFact includes all 300 development claims, each evaluated with three retrieved abstracts.
\begin{table}[t]
\caption{Study 2 populations. A family groups related claims and is resampled as one unit. SciFact's three abstracts for a claim remain together.}
\label{tab:data}
\begin{center}
\begin{tabular}{lrrrl}
\toprule
Dataset & Claims & Units & Families & Official metric \\
\midrule
FEVER & 2,000 & 2,000 & 1,963 & Strict score \\
FEVEROUS & 2,000 & 2,000 & 1,987 & Strict score \\
SciFact & 300 & 900 & 247 & Abstract-rationalized F1 \\
\bottomrule
\end{tabular}
\end{center}
\end{table}

We group claims into families using normalized exact matches, narrowly defined numeric/negation mutations, available citation links, and flagged near duplicates from each claim's top 50 lexical neighbors. The supplement records the assignments. This reduces known dependence, although semantic duplicates may remain. We froze the protocol internally after development and before the primary runs; its hashed record is included, without external preregistration.

\subsection{Evidence and Answer Interfaces}
Each evaluation unit has one candidate pool and two orderings: BM25 and \texttt{BAAI/bge-reranker-base}. For Wikipedia we retrieve five pages and retain at most 100 evidence units; for SciFact we retrieve three abstracts and rank their sentences. BGE reranks within these pools. We assemble whole evidence units in order until adding the next unit would exceed either model's evidence-token budget. Both models therefore see the same evidence text, with model-specific chat formatting. Appendix~\ref{app:configuration} gives retrieval queries, tie breaks, model revisions and truncation details.

The scorer receives a prefix of the visible evidence, limited to five FEVER sentences, five FEVEROUS sentences and 25 cells, or three rationale sentences per SciFact abstract. Thus increasing context can change both the model's input and the evidence submitted for scoring. Gold annotations are used for evaluation and prespecified controls; primary retrieval misses are retained. Gold SciFact abstracts outside the retrieved set remain in the relevant-abstract denominator.

We request either one code (A, B or C, mapped to the class labels) or one exact benchmark label string. After stripping whitespace, every other response is \invalid{}. We retain these responses and apply the specified scorer. Prompt-building functions, mappings and scorer code are public in the supplement; fully instantiated prompt records are archived separately.

The evidence rules differ across benchmarks. FEVER exempts correctly predicted NEI answers from evidence matching; the FEVEROUS implementation requires an annotated group even for NEI. SciFact's abstract-rationalized F1 is $2C/(P+R)$: $C$ counts correctly labeled and rationalized abstracts, $P$ submitted non-NEI abstracts, and $R$ relevant gold abstracts. A SciFact NEI or invalid answer omits that abstract from the submission. Invalid responses still count as incorrect in the separate answer-accuracy diagnostic, but their effect on the joint metric depends on its precision/recall denominator.

\subsection{Models and Controls}
We run \qwen{} and \llama{} with fixed weights in bfloat16 and a 16-token output limit. Qwen uses its non-thinking chat option. Each condition includes greedy decoding and five sampled replicates (seeds 1103, 2207, 3301, 4409 and 5519), with temperature 0.7, top-$p$ 0.8, top-$k$ 20 and repetition penalty 1.0. A deterministic hash of the dataset, claim, document and replicate sets each unit's seed, shared across sources, formats and budgets.

The primary matrix contains
\[
4{,}900\ \text{units}\times2\ \text{sources}\times2\ \text{contexts}
\times2\ \text{formats}\times2\ \text{models}\times6\ \text{replicates}
=470{,}400
\]
responses. Prespecified greedy controls test evidence removal, code mappings, cross-claim evidence swaps, complete annotated evidence where it fits, and evidence order. They require 42,834 additional generations after reuse of identical requests. Appendix~\ref{app:controls} reports each control's population, feasibility and results. All primary responses, including parsing failures, are saved; subsequent analyses reuse them.

\subsection{Inference and Decision Rule}
For each dataset--model pair, we average the format and context differences in $E_0$ over the other factor and five sampled seeds:
\begin{align}
D_{\mathrm{format}}&=\frac{1}{5}\sum_s\frac{1}{2}\sum_c
 [E_0(\mathrm{labels},c,s)-E_0(\mathrm{codes},c,s)],\\
D_{\mathrm{context}}&=\frac{1}{5}\sum_s\frac{1}{2}\sum_f
 [E_0(f,2048,s)-E_0(f,256,s)].
\end{align}
These form the 12 primary contrasts. Greedy outputs are descriptive references. We also partition the evidence gains by whether the two sources yield identical valid answers; for SciFact, all three abstract labels must match. Each partition keeps the relevant full-corpus denominator, so its contributions sum back to the original metric. We do not calculate a separate F1 for each subgroup.

For each panel, 50,000 bootstrap draws sample whole claim families and independently sample five decoding-seed indices with replacement. Each draw uses the same samples in every paired condition. We aggregate the scoring counts, compute the corpus metric for each sampled seed, then average. In particular, SciFact F1 is computed from corpus totals. The recorded implementation uses NumPy PCG64 seed 20260916 and batches of 128. Linear percentile bounds at $0.05/(2\times12)$ and $1-0.05/(2\times12)$ apply Bonferroni adjustment across the primary family. Coverage is approximate and conditional on the selected models and family construction.

The protocol calls an effect \emph{replicated dependence} when one factor has the same direction, magnitude of at least two score points, an adjusted interval excluding zero, and at least four of five agreeing seed signs for both models on the same two or more datasets, including SciFact. It calls all effects small (\emph{bounded stability}) when all 12 adjusted intervals lie inside $[-2,+2]$. All other outcomes receive the recorded label \emph{restricted/mixed evidence}. These labels describe prespecified decision rules, not new kinds of statistical evidence.

\section{LLM Results}
\subsection{Primary Contrasts}
\begin{figure}[t]
\centering\includegraphics[width=\textwidth]{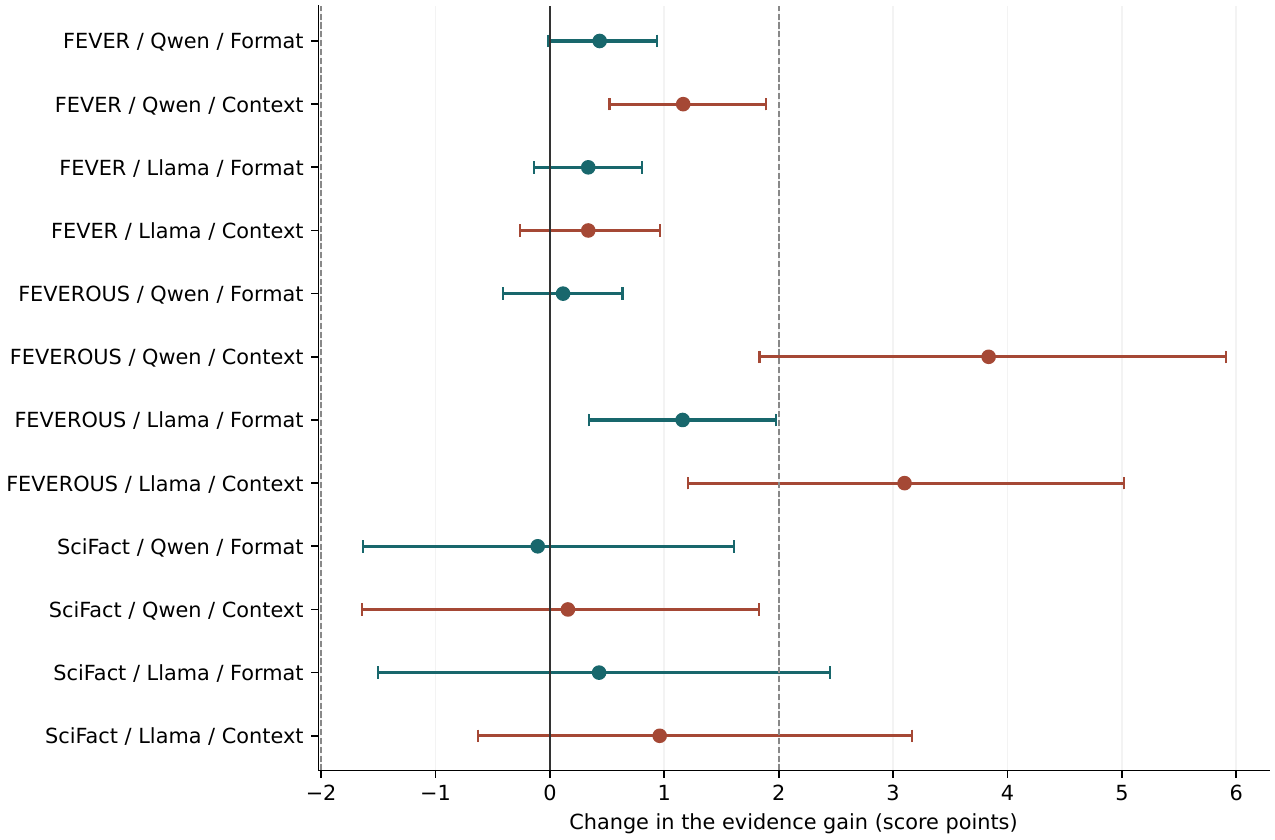}
\caption{Primary format and context effects on the fixed-answer evidence gain, with Bonferroni-adjusted intervals. Dashed lines mark the prespecified two-point threshold.}\label{fig:forest}
\end{figure}

Figure~\ref{fig:forest} shows the 12 primary contrasts; Appendix~\ref{app:primary} gives the exact values.
Increasing context raises $E_0$ on FEVEROUS by 3.84 points for Qwen [1.83, 5.91] and 3.10 for Llama [1.20, 5.01]; all five seed effects are positive. Both panels meet the local magnitude, interval and sign criteria. Neither factor, however, meets the required cross-dataset pattern. SciFact is required by that rule and qualifies for neither model; FEVER also fails the common two-point criterion. The same token increase retains different amounts of material on each dataset, as Section~\ref{sec:profiles} explains.

The smaller positive contrasts are Qwen's FEVER context effect (1.17, [0.52, 1.89]) and Llama's FEVEROUS format effect (1.16, [0.34, 1.98]). All other format intervals include zero. SciFact's adjusted context intervals are [-1.65, 1.83] for Qwen and [-0.63, 3.17] for Llama. The latter includes a FEVEROUS-sized effect, while the former excludes a positive two-point effect.

Several intervals extend beyond $[-2,+2]$. We can therefore establish neither the required replication nor that all effects lie within the small-effect bound. The recorded outcome is \emph{restricted/mixed evidence}: context dependence on FEVEROUS with inconclusive or smaller effects elsewhere.

\subsection{Endpoint Gains and Decomposition Order}
Evidence gain and endpoint performance can move in opposite directions. For Qwen on FEVER, increasing context raises $E_0$ by 1.17 points while the BGE endpoint $S_{11}$ falls by 1.83 points (Table~\ref{tab:budget}). The source gain $S_{11}-S_{00}$ increases by 1.65 points. On FEVEROUS, Qwen's $E_0$ rises from 2.28 to 6.11 while its BGE endpoint rises by 13.68. A gain from changing scored evidence, a gain from changing sources, and the performance at an endpoint answer different questions.
\begin{table}[ht]
\caption{The four scores and evidence gain by context budget, averaged over both answer formats and five sampled seeds. Scores use the same input and scoring budget within each row.}
\label{tab:budget}
\begin{center}
\begin{tabular}{llrrrrrr}
\toprule
Dataset & Model & Budget & $S_{00}$ & $S_{01}$ & $S_{10}$ & $S_{11}$ & $E_0$ \\
\midrule
FEVER & Qwen & 256 & 45.17 & 47.04 & 45.25 & 48.55 & 1.87 \\
FEVER & Qwen & 2,048 & 41.70 & 44.74 & 43.52 & 46.73 & 3.04 \\
FEVER & Llama & 256 & 38.33 & 40.24 & 37.14 & 39.98 & 1.91 \\
FEVER & Llama & 2,048 & 35.06 & 37.31 & 34.90 & 37.25 & 2.25 \\
FEVEROUS & Qwen & 256 & 15.72 & 18.00 & 15.40 & 19.88 & 2.28 \\
FEVEROUS & Qwen & 2,048 & 26.31 & 32.42 & 27.05 & 33.56 & 6.11 \\
FEVEROUS & Llama & 256 & 14.33 & 16.70 & 14.34 & 18.16 & 2.37 \\
FEVEROUS & Llama & 2,048 & 22.85 & 28.31 & 23.68 & 29.80 & 5.47 \\
SciFact & Qwen & 256 & 47.43 & 48.14 & 47.13 & 48.25 & 0.71 \\
SciFact & Qwen & 2,048 & 46.78 & 47.64 & 45.94 & 47.37 & 0.87 \\
SciFact & Llama & 256 & 42.57 & 43.87 & 42.17 & 43.56 & 1.30 \\
SciFact & Llama & 2,048 & 41.91 & 44.16 & 41.19 & 42.81 & 2.26 \\
\bottomrule
\end{tabular}
\end{center}
\end{table}

\begin{table}[ht]
\caption{Context effects on the two evidence gains and their average. Intervals use 1,000,000 draws with adjustment over the 60-comparison exploratory family. Full format and endpoint comparisons are in the supplement.}
\label{tab:orders}
\begin{center}
\begin{tabular}{llrll}
\toprule
Dataset & Model & $D_c E_0$ & $D_c E_1$ [interval] & $D_c E_{\mathrm{equal}}$ [interval] \\
\midrule
FEVER & Qwen & 1.17 & -0.10 [-0.61, 0.42] & 0.54 [0.01, 1.12] \\
FEVER & Llama & 0.34 & -0.49 [-1.14, 0.13] & -0.08 [-0.60, 0.46] \\
FEVEROUS & Qwen & 3.84 & 2.04 [-0.42, 4.52] & 2.94 [0.60, 5.31] \\
FEVEROUS & Llama & 3.10 & 2.30 [-0.07, 4.70] & 2.70 [0.47, 4.96] \\
SciFact & Qwen & 0.16 & 0.31 [-1.39, 2.66] & 0.23 [-0.58, 1.38] \\
SciFact & Llama & 0.96 & 0.23 [-1.76, 2.28] & 0.59 [-1.14, 2.60] \\
\bottomrule
\end{tabular}
\end{center}
\end{table}

The choice of fixed answers also matters. On FEVEROUS, Qwen's context contrast is 3.84 under $E_0$ and 2.04 under $E_1$; Llama's is 3.10 and 2.30. Averaging the two orders gives 2.94 and 2.70. Table~\ref{tab:orders} uses 1,000,000 paired bootstrap draws and adjustment over a 60-comparison exploratory family. The crossed-budget and direct interaction analyses below form a separate 42-comparison post-hoc family. These retain the original resampling units; their guarantees do not combine across families. Appendix~\ref{app:precision} records the numerical-precision checks.

For a concrete example, Qwen gives the same correct \texttt{SUPPORTS} answer for FEVEROUS claim 236 under both sources, but BGE omits a required sentence. The four scores are $(1,0,1,0)$. Claim 472 gives the reverse pattern, $(0,1,0,1)$. These post-hoc, lowest-ID examples illustrate how evidence alone can change credit; Appendix~\ref{app:examples} provides the traces and selection rules.

\subsection{Crossing Input and Scoring Budgets}
\label{sec:crossed}
A larger context budget can change both the generated answer and the evidence reaching the scorer. We examine their contributions by crossing the two budgets. Let $T_{ab}(u,v)$ score source-$a$ answers generated at input budget $u$ with source-$b$ evidence submitted at scoring budget $v$. The original conditions have $u=v$. Define $F(u,v)=T_{01}(u,v)-T_{00}(u,v)$, averaging the same formats and five per-seed corpus scores. For short and long budgets $s,l$,
\begin{align}
F(l,l)-F(s,s)
 &= [F(l,s)-F(s,s)] + [F(l,l)-F(l,s)] \\
 &= [F(s,l)-F(s,s)] + [F(l,l)-F(s,l)].
\end{align}
The first order changes input at fixed short scoring evidence, then scoring evidence at fixed long-input answers. The second reverses the steps. Both use the saved outputs.

On FEVEROUS, the scoring-budget component for Qwen is 2.91 points when input changes first, or 1.79 when scoring changes first, out of the 3.84-point context contrast. Llama's values are 2.56 and 2.23 out of 3.10. The remainder comes from changing the answers at fixed scoring evidence. For FEVER/Qwen, scoring changes contribute only 0.08 or 0.10 points, compared with input components of 1.07 and 1.09. SciFact has unchanged rationale eligibility at both budgets, making its scoring components and input--scoring interaction exactly zero for these submissions.

\begin{table}[htbp]
\caption{FEVEROUS context-effect decomposition in score points. Input-first holds scoring evidence at the short budget for the input change; scoring-first holds answers at the short input budget for the scoring change. Intervals use 1,000,000 draws and the 42-comparison adjustment. Appendix~\ref{app:paths} gives all datasets.}
\label{tab:path-ci}
\begin{center}
\begin{tabular}{lllrr}
\toprule
Dataset & Model & Path order & Input component [CI] & Scoring component [CI] \\
\midrule
FEVEROUS & Qwen & Input first & 0.93 [0.08, 1.87] & 2.91 [0.54, 5.28] \\
FEVEROUS & Qwen & Scoring first & 2.05 [0.85, 3.34] & 1.79 [-0.18, 3.79] \\
FEVEROUS & Llama & Input first & 0.55 [-0.38, 1.50] & 2.56 [0.42, 4.70] \\
FEVEROUS & Llama & Scoring first & 0.88 [-0.10, 1.89] & 2.23 [0.26, 4.21] \\
\bottomrule
\end{tabular}
\end{center}
\end{table}

The intervals use 1,000,000 paired draws and a separate 42-comparison post-hoc adjustment (tail probability $0.05/84$). The plan was recorded after inspecting earlier results and before these calculations. Full results for all datasets appear in Appendix~\ref{app:paths}.

\paragraph{What the intervals support.}
Llama's FEVEROUS scoring components exclude zero in both orders. Qwen's input components exclude zero in both orders; its scoring component does so only when scoring changes second. Input--scoring interaction intervals include zero in every panel. The direct context change in $E_1-E_0$ is -1.80 [-3.05, -0.64] for Qwen on FEVEROUS and -0.80 [-2.05, 0.41] for Llama (Table~\ref{tab:cpu-interactions}), supporting an order difference for Qwen. In the separate 60-comparison family, both FEVEROUS averaged-order intervals exclude zero and both $E_1$ intervals include it. The direct paired contrast, rather than these separate zero-exclusion outcomes, is the relevant comparison of orders.

\paragraph{Where the FEVER input component occurs.}
For FEVER/Qwen, BGE adds evidence eligibility on 104 claims at 256 tokens and 108 at 2,048; 103 are common to both sets. Their 109-claim union accounts arithmetically for the full input contribution. Within this post-hoc group, accuracy with BM25 evidence rises from 38.72\% to 59.54\%, while accuracy with BGE evidence changes from 64.04\% to 62.20\% (Table~\ref{tab:stratum}). The group's contributions, using the full 2,000-claim denominator, are 1.065 and 1.085 points under short and long scoring budgets; the remaining claims net to zero. This localizes the score change without requiring an account of the model's internal evidence use.
\begin{table}[htbp]
\caption{FEVER/Qwen answer accuracy in the post-hoc union of 109 claims where BGE adds eligibility: 104 at the short budget, 108 at the long budget and 103 at both. Means include both formats and five sampled seeds.}
\label{tab:stratum}
\begin{center}
\begin{tabular}{lrr}
\toprule
Input evidence & 256 tokens & 2,048 tokens \\
\midrule
BM25 & 38.72 & 59.54 \\
BGE & 64.04 & 62.20 \\
\bottomrule
\end{tabular}
\end{center}
\end{table}

\paragraph{Marginal changes miss the FEVER effect.}
Applying Equation~\ref{eq:covariance} at each budget reveals a contrast with Study~1. On FEVER/Qwen, the product of marginal answer accuracy and eligibility differences predicts a 0.110-point context change, compared with 1.165 observed. The changing covariance supplies 1.055 points. On FEVEROUS/Llama the same product predicts 3.068 points, close to the observed 3.100 (Table~\ref{tab:covariance}). Agreement for a change can occur even when the covariance at each budget is nonzero. The important question is which claims combine correct answers with improved eligibility.
\begin{table}[htbp]
\caption{Product and covariance contributions to the context effect in strict-score points. The last two terms of Equation~\ref{eq:covariance}, differenced between budgets, sum to the observed effect before rounding.}
\label{tab:covariance}
\begin{center}
\begin{tabular}{llrrr}
\toprule
Dataset & Model & Product change & Actual $D_cE_0$ & Covariance change \\
\midrule
FEVER & Qwen & 0.110 & 1.165 & 1.055 \\
FEVER & Llama & -0.010 & 0.335 & 0.345 \\
FEVEROUS & Qwen & 3.571 & 3.835 & 0.264 \\
FEVEROUS & Llama & 3.068 & 3.100 & 0.032 \\
\bottomrule
\end{tabular}
\end{center}
\end{table}

\subsection{Answer-Format Sensitivity}
The format contrast combines answer choice with compliance with the requested output format. On FEVEROUS, Llama produces at least one invalid response in 13.97\% of claim-condition observations with codes, compared with 8.16\% with label strings (Table~\ref{tab:invalid}). Each observation covers both sources; the rates pool both budgets and all six replicates. As a sensitivity check, we apply a conservative alternative parser to every saved response, accepting unambiguous surface variants without consulting gold labels. FEVEROUS context estimates remain positive and close to their original values. Appendix~\ref{app:parser} reports the remaining failures and definitions; Appendix~\ref{app:controls} gives the code-mapping and other controls.

\section{Why the Context Change Differs Across Datasets}
\label{sec:profiles}
The two budgets add different amounts of visible evidence across benchmarks. At 256 tokens, the average retained fraction is about 5\% on FEVEROUS, 30\% on FEVER and 65\% on SciFact; at 2,048 tokens it is about 44\%, 100\% and 100\%. Some material remains excluded at the larger budget for 1,998 of 2,000 FEVEROUS claims. Its structural categories include 739 mixed sentence/table claims and 463 table-or-cell claims. Thus the replication rule tests a common token-budget change, not an equal increase in information.

\begin{figure}[!htbp]
\centering\includegraphics[width=\textwidth]{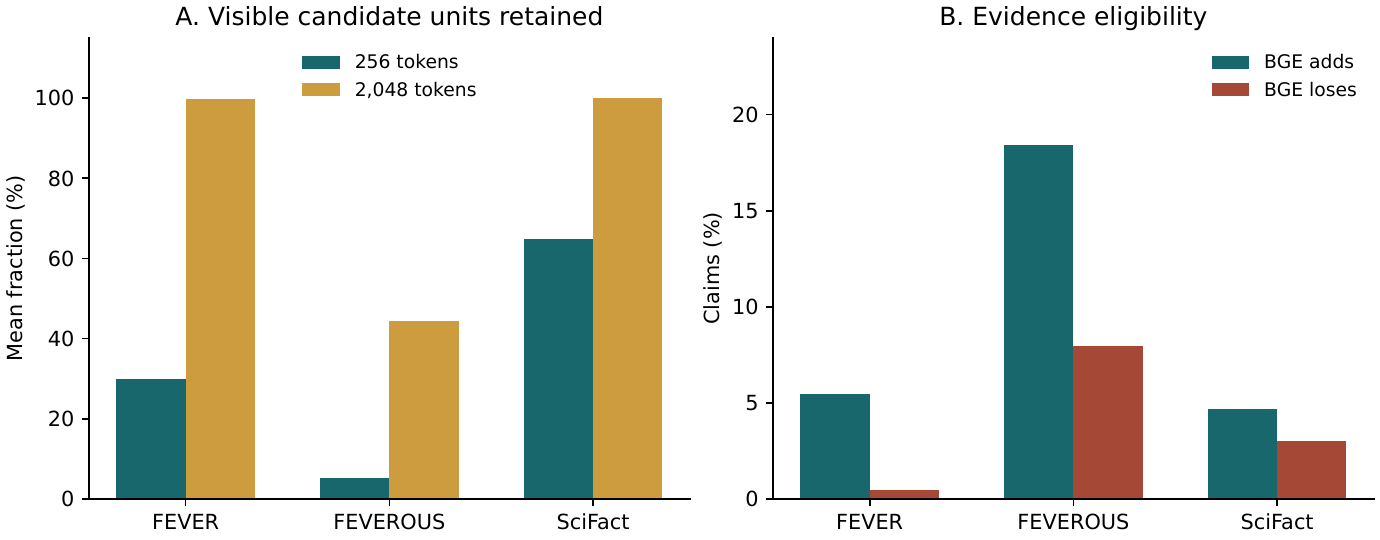}
\caption{Visible evidence retained and evidence eligibility under the official submission limits. A claim is in ``BGE adds'' if BGE improves eligibility at either budget and worsens it at neither; ``BGE loses'' reverses the comparison. Corrected profiles use normalized evidence identifiers (Appendix~\ref{app:correction}).}\label{fig:profiles}
\end{figure}

Figure~\ref{fig:profiles} also distinguishes visible evidence from eligible submissions. We corrected the FEVEROUS diagnostic to use normalized page identifiers and the evidence actually submitted within the scoring limits. This changes the profiles but leaves primary scores and intervals unchanged. SciFact additionally has 37 recorded document-retrieval misses among 300 claims. Its profile counts the eligible retrieved abstracts, and its corpus F1 differs from the two strict-accuracy metrics. These profiles characterize the different interventions; Appendix~\ref{app:correction} records the corrections and per-benchmark definitions.

\section{Discussion and Limitations}\label{sec:discussion}
\paragraph{What the two studies establish.}
In Study~1's DCUF-trained checkpoints, which predict no NEI, a 9.61-point joint-score gain contains 7.92 or 9.08 points from evidence replacement at fixed answers. In Study~2's two LLMs on FEVEROUS, this evidence gain changes with context budget. The studies provide complementary tests of the measurement question, but they use different sources, models and populations. Study~2 therefore leaves open whether the same interface dependence holds for Study~1's trained verifiers.

Both findings are consistent with the purpose of a joint metric: improving eligible evidence should improve the score. The added insight is how much of a particular gain remains without changing the answers, and which claims produce it. Marginal rates closely approximate Study~1's $E_0$, but the covariance term accounts for most of the FEVER/Qwen context contrast. Appendix~\ref{app:reproduce} describes the saved-output replays.

\paragraph{What to report.}
For a source comparison, retain each claim's answers and submitted evidence under both conditions. The official scorer can then produce the two off-diagonal scores without more inference. Report the four scores, both decomposition orders and answer accuracy, with class metrics and invalid-response rates where relevant. Distinguish model-visible evidence, submitted evidence and eligibility under the scoring rule. This requires access to both sets of outputs: an endpoint-only leaderboard cannot reconstruct the decomposition after the fact. For nonlinear metrics, retain the sufficient counts and recompute the corpus score.

\paragraph{Limits of Study 1.}
All checkpoints were trained on DCUF-derived inputs. Switching to UnifEE therefore also changes the input distribution, potentially affecting the answer components; the evidence share should not be extrapolated to source-symmetric or stronger verifiers. The 501 NEI-gold claims contribute zero to every cell, so the gains describe SUPPORTS/REFUTES performance on the full-population denominator. Source replacement also changes content, order, context, resolution and truncation together. The intervals condition on four fixed checkpoints. Saved predictions and historical reload records support the recorded comparison, while missing weights and runtime inputs limit full model reproduction. The retrospective override in Appendix~\ref{app:legacy} shares Study~1's claims and has unresolved threshold-selection history.

\paragraph{Limits of Study 2.}
We evaluate three English benchmarks, two fixed 8B models, two budgets and one reranker within fixed candidate pools. Decoding seeds do not capture training or model-selection uncertainty. The FEVEROUS panel comes from a historically exposed checkpoint-validation pool; residual semantic duplication and pretraining overlap are unknown. Evidence eligibility measures agreement with annotated identifiers, without new human relevance judgments. SciFact has 247 families and a different corpus metric; its 809 training claims lie outside the selected population.

The output-format sensitivity retains residual invalid responses and code-mapping effects. Many invalid Llama responses reach the output limit, as Appendix~\ref{app:revision} quantifies. The post-hoc budget and covariance analyses describe these saved outputs rather than model attention. Earlier source-matched LLM development gates failed; Appendix~\ref{app:development} distinguishes that development history from the completed study.

\section{Conclusion}
Joint fact-verification scores combine answer quality with evidence requirements. In four DCUF-trained verifiers with zero NEI recall, most of a 9.61-point source gain remains when answers are held fixed. In two LLMs on FEVEROUS, the fixed-answer evidence gain changes with context, while the prespecified cross-dataset criterion is unmet. The four answer--evidence score combinations expose these differences and reveal when aggregate accuracy and evidence-coverage rates are insufficient. They provide a reproducible account of what changed in a score, alongside the benchmark's intended endpoint measure.

\clearpage
\bibliographystyle{tmlr}
\bibliography{references}

\clearpage
\appendix
\section{Retrospective Case and Trained-Verifier Records}
\label{app:legacy}
\paragraph{Retrospective override.}
On 7,890 FEVEROUS development claims, an auxiliary four-model ensemble kept the public DCUF answer unless its mean NEI probability reached 0.23. Macro-F1 rose from \MayMacroD{} to \MayMacroH{} and NEI-F1 from zero to \MayNeiH{}, while accuracy fell from 72.05 to 69.01 and strict score from \MayStrictD{} to \MayStrictH{} (Table~\ref{tab:may-metrics}). Of 597 final NEI predictions, 595 changed and two stayed the same: 54 lose strict correctness, 24 gain it, and 519 remain incorrect. All 54 losses had eligible evidence according to the saved flags.
\begin{table}[htbp]
\caption{Retrospective NEI override: saved answer and strict-score metrics on 7,890 development claims.}
\label{tab:may-metrics}
\begin{center}
\begin{tabular}{lrrrr}
\toprule
Metric & DCUF & Hybrid & Difference & Stored 95\% interval \\
\midrule Macro-F1 & 49.34 & 54.16 & 4.82 & [3.67, 5.97] \\
Accuracy & 72.05 & 69.01 & -3.04 & [-3.56, -2.52] \\
NEI-F1 & 0.00 & 17.67 & 17.67 & [14.54, 20.77] \\
Strict & 35.16 & 34.78 & -0.38 & [-0.61, -0.16] \\
\bottomrule
\end{tabular}
\end{center}
\end{table}

We join the answers and per-model probabilities by FEVEROUS ID. Their mean reproduces the ensemble, and the 0.23 rule reproduces every hybrid answer. NEI is below the highest-probability class in 48 of the 54 losses; that class is the gold answer in 39. It is also below the highest-probability class in 17 of the 24 gains. The rule therefore both discards correct modal decisions and recovers gold NEI cases.

The retained records show development-informed candidate advancement and a later development sweep. They leave independent selection of the final threshold, four-model inclusion and rounded-mean rule unresolved. The original weights, successful-run source snapshot and exact model inputs are unavailable. This case is a retrospective description of the saved rule. It reuses seed numbers with Study~1, but uses different trained models.

\paragraph{Study 1 training and inputs.}
The base is \texttt{microsoft/deberta-large-mnli}, revision\
\texttt{7296194b9009373def4f7c5dad292651e4b5cf4e}. The class order is REFUTES, NEI, SUPPORTS. A label-independent ID hash partitions 71,291 training examples into 57,293 training, 7,002 checkpoint-validation and 6,996 reserved-calibration claims, all disjoint from the 7,890 development IDs. Each seed sees 16,000 distinct training examples. AdamW uses learning rate $10^{-5}$, weight decay 0.01, unweighted cross-entropy, FP32 microbatch 2 with accumulation 8, 60 warmup steps and decay to zero at update 1,000. Gradient clipping is 1.0; gradient checkpointing is enabled.

The input builder validates and deduplicates the first 30 candidate IDs, selects up to eight resolved units, and applies the block and sequence limits described in Section~\ref{sec:legacy}. Scoring preserves provider order within the 5/25 submission limits. Five DCUF and 235 UnifEE cases have unresolved direct units, including one shared case. The 7,651 both-source-resolved claims include 15 empty retrievals handled with the sentinel. The corpus audit checked archive-member CRCs and lengths; the published whole-corpus checksum was not independently reverified.

Historical reload records report unchanged checkpoint tensor hashes and zero reference-logit difference across source evaluations. The supplement contains these records, all eight prediction files and the paired labels/scoring flags. The trained weights and complete original inputs are excluded. Full model reproduction would require those missing assets.

\begin{table}[htbp]
\caption{Study 1 mean endpoint metrics and paired 95\% intervals for the 7,890-claim population. These are the original 10,000-draw claim-bootstrap results, conditional on the four checkpoints.}
\label{tab:legacy-metrics}
\begin{center}
\begin{tabular}{lrrrr}
\toprule
Metric & DCUF mean & UnifEE mean & Difference & Stored 95\% interval \\
\midrule Accuracy & 68.09 & 70.05 & 1.96 & [1.42, 2.50] \\
Macro-F1 & 46.14 & 47.58 & 1.44 & [1.03, 1.84] \\
NEI-F1 & 0.00 & 0.00 & 0.00 & [0.00, 0.00] \\
Strict & 32.41 & 42.02 & 9.61 & [8.77, 10.43] \\
\bottomrule
\end{tabular}
\end{center}
\end{table}

\begin{table}[htbp]
\caption{Study 1 endpoint metrics by training seed. Each checkpoint is evaluated on both evidence sources.}
\label{tab:legacy-seeds}
\begin{center}
\begin{tabular}{lrrrrr}
\toprule
Training seed & $S_{00}$ & $S_{01}$ & $S_{10}$ & $S_{11}$ & $G$ \\
\midrule 97113 & 32.64 & 40.49 & 32.85 & 41.98 & 9.34 \\
97117 & 31.98 & 39.82 & 32.94 & 41.90 & 9.92 \\
97119 & 32.31 & 40.33 & 32.78 & 41.80 & 9.49 \\
97121 & 32.71 & 40.68 & 33.19 & 42.40 & 9.68 \\
\bottomrule
\end{tabular}
\end{center}
\end{table}

\begin{table}[htbp]
\caption{Study 1 contributions by gold label. Each contribution uses the full population of 7,890 claims as its denominator; columns sum to the corresponding overall scores before rounding. NEI contributes zero because every checkpoint predicts only SUPPORTS or REFUTES.}
\label{tab:study1-classes}
\begin{center}
\begin{tabular}{lrrrrrr}
\toprule
Gold label & Claims & $S_{00}$ & $S_{01}$ & $S_{10}$ & $S_{11}$ & Gain $G$ \\
\midrule
SUPPORTS & 3908 & 19.61 & 25.01 & 19.84 & 25.54 & 5.93 \\
REFUTES & 3481 & 12.80 & 15.33 & 13.10 & 16.48 & 3.69 \\
NEI & 501 & 0.00 & 0.00 & 0.00 & 0.00 & 0.00 \\
\bottomrule
\end{tabular}
\end{center}
\end{table}

\paragraph{Interval reproduction and additional summaries.}
The original metric implementation reconstructs all 180 stored interval rows: 18 endpoint metrics for each of four checkpoints and their mean, in both the 7,890-claim and 7,651-claim populations. For each population, it resets PCG64 to 20261099, draws 10,000 paired claim samples, computes each checkpoint's source difference and averages. Linear 2.5th and 97.5th percentiles give pointwise intervals. Undefined ratios remain undefined. The stored CSVs and original implementation are unchanged.

A separate post-hoc script applies the same primary-population draws to $E_0,A_1,A_0,E_1$ and $I$. It reproduces the original total-gain interval as a cross-check and reports pointwise 95\% intervals for the five new summaries. These describe sampling variation conditional on the four checkpoints, separately from all LLM adjustment families. The script also gives the class contributions in Table~\ref{tab:study1-classes} and the product/covariance terms for both evidence gains. The original May intervals remain archived results, outside this replay.

\section{Development History and Study Populations}
\label{app:development}
An earlier LLM development study used DCUF/UnifEE evidence and two input contexts on 2,048 claims. Both model gates failed, as did a revised Llama label interface. That study stopped before its planned primary evaluation. Six exploratory scoring blocks can be reconstructed from 24,576 decisions, with bounds retained for one parser-unsupported evidence ID. The completed BM25/BGE study uses different candidates and conditions under its own later protocol.

ID joins show 7,890 shared claims between the retrospective override and Study~1. The final FEVEROUS LLM panel overlaps neither Study~1's development IDs nor the early LLM panel's IDs. Both LLM panels come from the historical 7,002-claim checkpoint-validation split. The supplement includes the membership projection and join procedure. This establishes ID disjointness, while the historical exposure, semantic-duplicate and pretraining limitations remain as described in Section~\ref{sec:discussion}. Detailed execution history and unexecuted plans are retained in the supplementary method notes.

\section{LLM Configuration and Evidence Construction}
\label{app:configuration}
Qwen uses \texttt{Qwen/Qwen3-8B}, revision\
\texttt{b968826d9c46dd6066d109eabc6255188de91218}; Llama uses\
\texttt{meta-llama/Llama-3.1-8B-Instruct}, revision\
\texttt{0e9e39f249a16976918f6564b8830bc894c89659}. Tokenizer and weight-file hashes are in the protocol. We process one request at a time. The shared evidence text fits both tokenizers; each model applies its own chat serialization and the recorded per-unit seed.

For FEVER and FEVEROUS, A/B/C map to SUPPORTS/REFUTES/NOT ENOUGH INFO. For SciFact they map to SUPPORT/CONTRADICT/NOT\_ENOUGH\_INFO. After whitespace stripping, a response must equal one code or the appropriate benchmark label. Other strings are invalid. FEVEROUS page IDs use Unicode NFD normalization for scoring; display text and raw outputs are retained unchanged.

Wikipedia candidates come from SQLite FTS5 BM25 retrieval over titles and bodies. Queries are the OR of unique casefolded alphanumeric claim tokens. We retrieve five pages, rank their evidence units lexically, and keep at most 100, with ties broken by evidence ID. SciFact retrieves three abstracts using the same engine and ranks at most 100 sentences per abstract. BGE reranks within these pools at revision \texttt{af37ed791788201a1cdcf513e0f584f3aa3be105}; reranker pairs are limited to 512 tokens and truncation is recorded. Material assembly stops when the next whole unit would exceed either tokenizer's budget. FEVEROUS cells retain their recorded table context and identities.

The original protocol specifies IDs, families, materials, parsers, models, requests, analysis code and compute limits. Its statistical code and outputs are unchanged. Reference code contains the source-view, candidate and prompt-construction policies, but the compact archive excludes the full corpora and a self-contained indexing pipeline. Fully instantiated prompts and operational logs remain in the private execution archive.

Input preparation excludes one FEVEROUS source record with an empty page identifier; the annotation audit found no references to such pages. FEVER has recorded repairs to malformed hyperlink metadata that preserve sentence IDs and text. Compatibility and packaging changes, including the local analysis timeout, are documented in the supplementary method notes.

\section{Complete Primary Results}
\label{app:primary}
\begin{table}[htbp]
\caption{All 12 primary effects in score points. Format is labels minus codes; context is 2,048 minus 256 tokens. Seed signs use each unrounded effect and the direction of the mean.}
\label{tab:primary}
\begin{center}
\begin{tabular}{lllrrr}
\toprule
Dataset & Model & Factor & Effect & Adjusted interval & Seed signs \\
\midrule
FEVER & Qwen & Format & 0.44 & [-0.02, 0.94] & 5/5 \\
FEVER & Qwen & Context & 1.17 & [0.52, 1.89] & 5/5 \\
FEVER & Llama & Format & 0.34 & [-0.14, 0.81] & 5/5 \\
FEVER & Llama & Context & 0.34 & [-0.26, 0.96] & 5/5 \\
FEVEROUS & Qwen & Format & 0.12 & [-0.41, 0.63] & 5/5 \\
FEVEROUS & Qwen & Context & 3.84 & [1.83, 5.91] & 5/5 \\
FEVEROUS & Llama & Format & 1.16 & [0.34, 1.98] & 5/5 \\
FEVEROUS & Llama & Context & 3.10 & [1.20, 5.01] & 5/5 \\
SciFact & Qwen & Format & -0.11 & [-1.64, 1.61] & 4/5 \\
SciFact & Qwen & Context & 0.16 & [-1.65, 1.83] & 3/5 \\
SciFact & Llama & Format & 0.43 & [-1.50, 2.45] & 4/5 \\
SciFact & Llama & Context & 0.96 & [-0.63, 3.17] & 5/5 \\
\bottomrule
\end{tabular}
\end{center}
\end{table}

Table~\ref{tab:primary} gives every prespecified contrast shown in Figure~\ref{fig:forest}.

\section{Per-Seed Results}
\begin{table}[htbp]
\caption{Primary effects by sampled decoding seed, in score points. Displayed zeros may be small nonzero values before rounding.}
\label{tab:seeds}
\begin{center}
\begin{tabular}{lllrrrrr}
\toprule
Dataset & Model & Factor & 1103 & 2207 & 3301 & 4409 & 5519 \\
\midrule
FEVER & Qwen & Format & 0.43 & 0.48 & 0.40 & 0.45 & 0.43 \\
FEVER & Qwen & Context & 1.18 & 1.18 & 1.15 & 1.15 & 1.18 \\
FEVER & Llama & Format & 0.10 & 0.53 & 0.33 & 0.40 & 0.33 \\
FEVER & Llama & Context & 0.40 & 0.13 & 0.23 & 0.45 & 0.48 \\
FEVEROUS & Qwen & Format & 0.10 & 0.13 & 0.13 & 0.13 & 0.10 \\
FEVEROUS & Qwen & Context & 3.90 & 3.83 & 3.93 & 3.78 & 3.75 \\
FEVEROUS & Llama & Format & 1.40 & 1.23 & 1.33 & 0.68 & 1.18 \\
FEVEROUS & Llama & Context & 2.75 & 3.08 & 3.38 & 3.13 & 3.18 \\
SciFact & Qwen & Format & -0.25 & -0.01 & -0.02 & -0.26 & 0.00 \\
SciFact & Qwen & Context & -0.29 & 0.46 & 0.46 & 0.20 & -0.04 \\
SciFact & Llama & Format & 0.45 & 1.17 & 0.10 & -0.02 & 0.45 \\
SciFact & Llama & Context & 0.85 & 1.13 & 0.60 & 0.88 & 1.35 \\
\bottomrule
\end{tabular}
\end{center}
\end{table}

Each entry uses one sampled replicate's corpus score. Sign counts use unrounded values. Greedy results and all 48 condition-level score tables are in \texttt{tables/native-score-table.csv} in the LLM supplement.

\section{Control Populations and Results}
\label{app:controls}
Controls use 64 deterministic hash-selected claims per dataset, expanded to three abstracts per SciFact claim where required. They remove evidence, permute the code mapping, swap evidence between eligible claims, reorder evidence units, or supply a complete annotated group that fits the budget. The controls are descriptive and were not used to retune primary conditions.
\begin{table}[htbp]
\caption{Planned and feasible control evaluations and the number of additional requests. Identical requests are reused; 56 SciFact requests reuse primary outputs.}
\label{tab:controls}
\begin{center}
\begin{tabular}{llrrrr}
\toprule
Dataset & Model & Planned & Feasible & Infeasible & New requests \\
\midrule
FEVER & Qwen & 4,736 & 4,544 & 192 & 4,473 \\
FEVER & Llama & 4,736 & 4,544 & 192 & 4,473 \\
FEVEROUS & Qwen & 4,736 & 4,548 & 188 & 4,484 \\
FEVEROUS & Llama & 4,736 & 4,548 & 188 & 4,484 \\
SciFact & Qwen & 13,952 & 13,212 & 740 & 12,460 \\
SciFact & Llama & 13,952 & 13,212 & 740 & 12,460 \\
\bottomrule
\end{tabular}
\end{center}
\end{table}

Of 46,848 planned control evaluations, 44,608 are feasible: 1,848 lack a fitting annotated group and 392 lack five distinct nonidentity orders. Repeated identical evaluations share requests, leaving 42,890 unique requests. Of these, 56 reuse primary outputs and 42,834 require additional generation. The case-level export records feasibility, answer correctness and agreement with the canonical control.
\begin{table}[htbp]
\caption{Descriptive control results (percent). Claim-only and Gold report answer accuracy on their respective feasible subsets. Mapping, Swap and Order report agreement with the corresponding canonical control. Full counts are in the supplement.}
\label{tab:control-diagnostics}
\begin{center}
\begin{tabular}{llrrrrr}
\toprule
Dataset & Model & Claim-only & Gold & Mapping & Swap & Order \\
\midrule
FEVER & Qwen & 39.1 & 78.8 & 84.8 & 53.9 & 82.1 \\
FEVER & Llama & 24.2 & 44.2 & 52.7 & 53.5 & 80.7 \\
FEVEROUS & Qwen & 1.6 & 79.7 & 85.9 & 36.5 & 85.4 \\
FEVEROUS & Llama & 1.6 & 73.4 & 69.8 & 24.4 & 79.6 \\
SciFact & Qwen & 83.9 & 67.7 & 93.3 & 78.5 & 93.8 \\
SciFact & Llama & 79.2 & 67.7 & 69.2 & 80.3 & 91.4 \\
\bottomrule
\end{tabular}
\end{center}
\end{table}

Qwen returns NEI for every claim-only evaluation; Llama returns NEI or an invalid answer in all but one. These prompts request evidence-grounded answers and remove the evidence, so the result is consistent with abstention. It does not measure how much relevant knowledge the model stores. Code-mapping disagreement likewise identifies interface sensitivity without isolating its cause.

\section{Examples with Unchanged Answers}
\label{app:examples}
For FEVEROUS claim 236, ``Mizmaar released a second album in 2007 which launched worldwide and two of the tracks became instant hits,'' Qwen returns \texttt{SUPPORTS} under both sources in the greedy, 2,048-token label condition. The annotated group contains \texttt{Mizmaar} sentences 4, 18, 19 and 20. BM25 submits [13, 4, 19, 18, 20]; BGE submits [4, 20, 18, 13] and \texttt{Kashan Admani} sentence 14. BGE omits required sentence 19, giving per-claim scores $(1,0,1,0)$. Claim 472 has unchanged valid answers and the reverse scores $(0,1,0,1)$. We selected the lowest-ID eligible example in each direction post hoc. Full evidence lists and responses accompany the records.
\begin{table}[htbp]
\caption{Twelve selected greedy examples. Entries are changes in the credited numerator when scored evidence changes. C/L denote codes/labels; Ex./Ctr. denote example/counterexample.}
\label{tab:examples}
\begin{center}
\begin{tabular}{llllrrrrr}
\toprule
Dataset & Model & Factor & Role & Claim ID & C/256 & C/2048 & L/256 & L/2048 \\
\midrule
FEVER & Qwen & Fmt & Ex. & 104617 & +0 & +0 & +1 & +1 \\
FEVER & Llama & Fmt & Ctr. & 18179 & +0 & +0 & +0 & +0 \\
FEVER & Llama & Ctx & Ex. & 84799 & +0 & +1 & +0 & +1 \\
FEVER & Qwen & Ctx & Ctr. & 200259 & +0 & +0 & +0 & +0 \\
FEVEROUS & Llama & Fmt & Ex. & 25328 & +0 & +0 & +1 & +1 \\
FEVEROUS & Qwen & Fmt & Ctr. & 33371 & +0 & +0 & +0 & +0 \\
FEVEROUS & Qwen & Ctx & Ex. & 86151 & +0 & +1 & +0 & +1 \\
FEVEROUS & Llama & Ctx & Ctr. & 68759 & +0 & +0 & +0 & +0 \\
SciFact & Qwen & Fmt & Ex. & 129 & -2 & +0 & -2 & -2 \\
SciFact & Llama & Fmt & Ctr. & 1107 & +0 & +0 & +0 & +0 \\
SciFact & Llama & Ctx & Ex. & 5 & +2 & +0 & +2 & +0 \\
SciFact & Qwen & Ctx & Ctr. & 1278 & +0 & +0 & +0 & +0 \\
\bottomrule
\end{tabular}
\end{center}
\end{table}

Table~\ref{tab:examples} retains the twelve examples selected under the earlier fixed policy. Each cell is a change in the credited numerator under greedy decoding when BM25 answers are scored with BGE instead of BM25 evidence. Code-to-label contrasts subtract the two code entries from the two label entries; context contrasts subtract the two short-budget entries from the two long-budget entries. The units are numerator counts, before conversion to corpus-score points. The supplement includes all traces and record hashes.

\section{Secondary Answer and Evidence Diagnostics}
\label{app:revision}
\begin{table}[htbp]
\caption{Budget-averaged score decomposition and the contribution from identical valid answers, in points. Partition contributions use the full-corpus denominator.}
\label{tab:accounting}
\begin{center}
\begin{tabular}{llrrrrrrr}
\toprule
Dataset & Model & $S_{00}$ & $S_{11}$ & $G$ & $E_0$ & $E_1$ & $I$ & $E_{0,U}$ \\
\midrule
FEVER & Qwen & 43.44 & 47.64 & 4.20 & 2.45 & 3.26 & 0.81 & 2.40 \\
FEVER & Llama & 36.70 & 38.61 & 1.92 & 2.08 & 2.59 & 0.52 & 1.86 \\
FEVEROUS & Qwen & 21.01 & 26.72 & 5.70 & 4.19 & 5.49 & 1.30 & 4.45 \\
FEVEROUS & Llama & 18.59 & 23.98 & 5.39 & 3.92 & 4.97 & 1.06 & 3.84 \\
SciFact & Qwen & 47.10 & 47.81 & 0.71 & 0.79 & 1.27 & 0.48 & 0.75 \\
SciFact & Llama & 42.24 & 43.19 & 0.95 & 1.78 & 1.51 & -0.27 & 1.24 \\
\bottomrule
\end{tabular}
\end{center}
\end{table}

The unchanged-answer contribution in Table~\ref{tab:accounting} uses the full-corpus denominator. This lets contributions from the partitions sum to the total evidence gain, including for corpus F1. Negative answer components can offset positive evidence components.
\begin{table}[htbp]
\caption{Individual responses, including greedy. Surface variants and unambiguous leading-label explanations are accepted by the alternative parser; Remaining covers its residual failures. The final column counts originally invalid responses with 16 tokens and overlaps the other categories.}
\label{tab:taxonomy}
\begin{center}
\begin{tabular}{lllrrrrrr}
\toprule
Dataset & Model & Format & $N$ & Invalid & Surface & Leading & Remaining & 16 tokens \\
\midrule
FEVER & Qwen & codes & 48000 & 4 & 0 & 4 & 0 & 4 \\
FEVER & Qwen & labels & 48000 & 0 & 0 & 0 & 0 & 0 \\
FEVER & Llama & codes & 48000 & 2080 & 101 & 639 & 1340 & 1939 \\
FEVER & Llama & labels & 48000 & 2515 & 46 & 1649 & 820 & 2462 \\
FEVEROUS & Qwen & codes & 48000 & 29 & 0 & 29 & 0 & 29 \\
FEVEROUS & Qwen & labels & 48000 & 0 & 0 & 0 & 0 & 0 \\
FEVEROUS & Llama & codes & 48000 & 4481 & 402 & 790 & 3289 & 4026 \\
FEVEROUS & Llama & labels & 48000 & 2379 & 43 & 1500 & 836 & 2335 \\
SciFact & Qwen & codes & 21600 & 10 & 0 & 9 & 1 & 10 \\
SciFact & Qwen & labels & 21600 & 0 & 0 & 0 & 0 & 0 \\
SciFact & Llama & codes & 21600 & 122 & 4 & 95 & 23 & 113 \\
SciFact & Llama & labels & 21600 & 843 & 0 & 718 & 125 & 843 \\
\bottomrule
\end{tabular}
\end{center}
\end{table}

\begin{table}[htbp]
\caption{Invalid Llama responses truncated at the generation limit: 16 tokens with no terminal token (IDs 128001, 128008 or 128009). The percentage uses invalid responses as its denominator. Both sampled-only and including-greedy results are shown.}
\label{tab:cap}
\begin{center}
\begin{tabular}{llrrr}
\toprule
Dataset & Population & Responses & Truncated / invalid & Rate \\
\midrule
FEVER & Including greedy & 96,000 & 4,389/4,595 & 95.52\% \\
FEVER & Sampled only & 80,000 & 3,846/4,036 & 95.29\% \\
FEVEROUS & Including greedy & 96,000 & 6,359/6,860 & 92.70\% \\
FEVEROUS & Sampled only & 80,000 & 5,451/5,909 & 92.25\% \\
SciFact & Including greedy & 43,200 & 955/965 & 98.96\% \\
SciFact & Sampled only & 36,000 & 829/837 & 99.04\% \\
\bottomrule
\end{tabular}
\end{center}
\end{table}

The Llama configuration sets a 16-token output limit and terminal IDs 128001, 128008 and 128009, without custom stopping criteria or strings. It saves the generated token slice before removing special tokens. A 16-token output with no terminal token is therefore truncated at the generation limit. Table~\ref{tab:cap} identifies that subset, which is slightly smaller than the broad 16-token count in Table~\ref{tab:taxonomy}. Many responses already contain explanations that violate the exact-label rule. The saved records establish truncation, but whether longer completions would become valid or more accurate requires a different experiment.

\begin{table}[htbp]
\caption{Answer accuracy ($L$), joint score with oracle-correct answers ($O$), and observed joint score ($S$). Subscripts 0/1 denote BM25/BGE. SciFact accuracy uses 900 claim--abstract pairs; its $O$ and $S$ are corpus F1, including gold abstracts missed by retrieval.}
\label{tab:label-eligibility}
\begin{center}
\begin{tabular}{llrrrrrrr}
\toprule
Dataset & Model & Budget & $L_0$ & $L_1$ & $O_0$ & $O_1$ & $S_{00}$ & $S_{11}$ \\
\midrule
FEVER & Qwen & 256 & 57.44 & 60.38 & 65.35 & 70.10 & 45.17 & 48.55 \\
FEVER & Qwen & 2,048 & 56.78 & 59.38 & 65.40 & 70.40 & 41.70 & 46.73 \\
FEVER & Llama & 256 & 51.01 & 52.76 & 65.35 & 70.10 & 38.33 & 39.98 \\
FEVER & Llama & 2,048 & 48.25 & 48.88 & 65.40 & 70.40 & 35.06 & 37.25 \\
FEVEROUS & Qwen & 256 & 41.20 & 45.96 & 26.00 & 29.85 & 15.72 & 19.88 \\
FEVEROUS & Qwen & 2,048 & 56.37 & 57.55 & 41.25 & 50.40 & 26.31 & 33.56 \\
FEVEROUS & Llama & 256 & 44.09 & 50.35 & 26.00 & 29.85 & 14.33 & 18.16 \\
FEVEROUS & Llama & 2,048 & 52.08 & 54.33 & 41.25 & 50.40 & 22.85 & 29.80 \\
SciFact & Qwen & 256 & 87.10 & 87.12 & 72.43 & 75.14 & 47.43 & 48.25 \\
SciFact & Qwen & 2,048 & 85.97 & 86.37 & 72.43 & 75.14 & 46.78 & 47.37 \\
SciFact & Llama & 256 & 82.16 & 82.26 & 72.43 & 75.14 & 42.57 & 43.56 \\
SciFact & Llama & 2,048 & 82.81 & 82.79 & 72.43 & 75.14 & 41.91 & 42.81 \\
\bottomrule
\end{tabular}
\end{center}
\end{table}

\begin{table}[htbp]
\caption{Direct paired interaction estimates with intervals adjusted over the 42-comparison post-hoc family. The first column compares the two input-budget components; the others compare evidence-decomposition orders. SciFact\textquotesingle s input--scoring interaction is zero by construction because rationale eligibility is unchanged across budgets at fixed answers.}
\label{tab:cpu-interactions}
\begin{center}
\begin{tabular}{llrrr}
\toprule
Dataset & Model & Input $\times$ scoring & $D_c(E_1-E_0)$ & $D_f(E_1-E_0)$ \\
\midrule
FEVER & Qwen & 0.02 [-0.23, 0.22] & -1.26 [-2.06, -0.59] & 0.29 [-0.28, 0.90] \\
FEVER & Llama & 0.03 [-0.15, 0.20] & -0.82 [-1.66, -0.04] & 0.02 [-0.52, 0.57] \\
FEVEROUS & Qwen & 1.12 [-0.23, 2.51] & -1.80 [-3.05, -0.64] & 0.26 [-0.35, 0.88] \\
FEVEROUS & Llama & 0.33 [-0.78, 1.43] & -0.80 [-2.05, 0.41] & -0.58 [-1.48, 0.36] \\
SciFact & Qwen & 0.00 [0.00, 0.00] & 0.15 [-3.02, 3.85] & 0.04 [-1.43, 1.48] \\
SciFact & Llama & 0.00 [0.00, 0.00] & -0.73 [-2.94, 0.79] & -0.60 [-2.78, 0.92] \\
\bottomrule
\end{tabular}
\end{center}
\end{table}

The secondary records include the original and corrected coverage diagnostics, parser sensitivity and crossed-budget analysis. The later 42-comparison plan was recorded after earlier results were inspected and before its own calculations. Original results, exact claim IDs, covariance terms and Monte Carlo batches are retained with their separate analysis plans.

\section{Alternative Parser and Invalid-Response Rates}
\label{app:parser}
\begin{table}[htbp]
\caption{Claim-condition observations flagged for any invalid response, by answer format. Each observation covers both sources and, for SciFact, all three abstracts. Rates pool both budgets and six replicates.}
\label{tab:invalid}
\begin{center}
\begin{tabular}{lllrr}
\toprule
Dataset & Model & Format & Any invalid / observations & Rate \\
\midrule
FEVER & Qwen & codes & 4/24,000 & 0.02\% \\
FEVER & Qwen & labels & 0/24,000 & 0.00\% \\
FEVER & Llama & codes & 1,565/24,000 & 6.52\% \\
FEVER & Llama & labels & 1,980/24,000 & 8.25\% \\
FEVEROUS & Qwen & codes & 24/24,000 & 0.10\% \\
FEVEROUS & Qwen & labels & 0/24,000 & 0.00\% \\
FEVEROUS & Llama & codes & 3,353/24,000 & 13.97\% \\
FEVEROUS & Llama & labels & 1,959/24,000 & 8.16\% \\
SciFact & Qwen & codes & 7/3,600 & 0.19\% \\
SciFact & Qwen & labels & 0/3,600 & 0.00\% \\
SciFact & Llama & codes & 85/3,600 & 2.36\% \\
SciFact & Llama & labels & 536/3,600 & 14.89\% \\
\bottomrule
\end{tabular}
\end{center}
\end{table}

Table~\ref{tab:invalid} flags a claim-condition observation when either source's answer is invalid. For SciFact, any invalid answer among two sources and three abstracts triggers the flag. The denominator includes both contexts and six replicates for each format. A claim may therefore contribute several flagged observations.

Qwen rarely fails the exact-label format. Llama's failures vary with format: on FEVEROUS, 13.97\% of code observations and 8.16\% of label observations are flagged. This interface behavior is part of the primary format comparison.
\begin{table}[htbp]
\caption{Original and alternative-parser effects in score points. The alternative parser accepts unambiguous variants without consulting gold labels and keeps every observation in the original population.}
\label{tab:parser}
\begin{center}
\begin{tabular}{llrrrr}
\toprule
Dataset & Model & Format strict & Format alt. & Context strict & Context alt. \\
\midrule
FEVER & Qwen & 0.44 & 0.44 & 1.17 & 1.17 \\
FEVER & Llama & 0.34 & 0.27 & 0.34 & 0.36 \\
FEVEROUS & Qwen & 0.12 & 0.12 & 3.84 & 3.84 \\
FEVEROUS & Llama & 1.16 & 1.04 & 3.10 & 3.27 \\
SciFact & Qwen & -0.11 & -0.11 & 0.16 & 0.16 \\
SciFact & Llama & 0.43 & 0.44 & 0.96 & 0.95 \\
\bottomrule
\end{tabular}
\end{center}
\end{table}

The alternative parser examines every response without consulting its gold label. It normalizes case, spaces/underscores and surrounding punctuation, and accepts a leading label/code followed by an explanation only when there is one unambiguous interpretation. Conflicting labels, uncertainty or negation markers, questions, empty strings and other malformed outputs remain invalid. The FEVEROUS context estimates remain close to their original values. Residual failures and mapping effects still limit a parser-independent interpretation.

\section{Evidence-Eligibility Correction}
\label{app:correction}
The original FEVEROUS diagnostic used raw page IDs and visible evidence. The corrected version uses the scorer's NFD-normalized IDs and the submitted prefix. Normalization changes visible-group membership in 210 of 8,000 presentations. In 497 presentations a complete normalized group is visible but falls outside the submitted prefix.

Corrected evidence eligibility increases under BGE for 368 FEVEROUS claims and decreases for 159; the corresponding counts are 109/9 for FEVER and 14/9 for SciFact. A gain means improvement at one or both budgets and deterioration at neither; a loss reverses the definition. FEVER correct NEI answers require no evidence match, so their eligibility is one even when annotated group coverage is undefined. SciFact compares the number of eligible retrieved abstracts per claim; abstracts without annotated rationales have undefined group coverage.

The correction resolves 52 Qwen and 59 Llama claims whose strict successes conflicted with the old flags. For example, claim 837 has a complete normalized submitted group at 2,048 tokens under both sources. Re-parsing and rescoring the raw responses reproduces all original counts, with zero strict successes lacking corrected eligibility. Original diagnostic files and the correction are both retained; the primary effects and intervals are unchanged.

\section{Secondary Intervals and Monte Carlo Precision}
\label{app:precision}
The primary family uses 50,000 draws and tail probability $0.05/24$, about 104 expected draws per tail. The 60-comparison exploratory family uses tail probability $0.05/120$: increasing its draws from 50,000 to 1,000,000 raises that count from about 21 to 417. The separate 42-comparison family uses $0.05/84$, about 595 draws per tail at 1,000,000 draws. These counts describe numerical resolution of the percentile bounds, not additional experimental observations. The original primary analysis is retained.

The million-draw results pool four independently seeded 250,000-draw batches, specified before that calculation. For Llama on FEVEROUS, the $E_1$ context lower bound is -0.075; batch lower bounds range from -0.079 to -0.065. The bound is close to zero. Table~\ref{tab:mc} reports batch endpoints, while direct order comparisons appear in Table~\ref{tab:cpu-interactions}.
\begin{table}[htbp]
\caption{Lower and upper percentile endpoints for predetermined 250,000-draw batches at the 60-comparison tail probabilities. The reported intervals pool all four batches. All 102 secondary contrast records are retained in the supplement.}
\label{tab:mc}
\begin{center}
\begin{tabular}{llrrr}
\toprule
Dataset & Model & PCG64 stream & $E_1$ context lower & Upper \\
\midrule
FEVEROUS & Qwen & 202609201 & -0.426 & 4.581 \\
FEVEROUS & Qwen & 202609202 & -0.415 & 4.489 \\
FEVEROUS & Qwen & 202609203 & -0.420 & 4.521 \\
FEVEROUS & Qwen & 202609204 & -0.394 & 4.523 \\
FEVEROUS & Llama & 202609201 & -0.065 & 4.675 \\
FEVEROUS & Llama & 202609202 & -0.075 & 4.706 \\
FEVEROUS & Llama & 202609203 & -0.070 & 4.710 \\
FEVEROUS & Llama & 202609204 & -0.079 & 4.678 \\
\bottomrule
\end{tabular}
\end{center}
\end{table}

\section{Complete Crossed-Budget Results}
\label{app:paths}
\begin{table}[htbp]
\caption{All crossed-budget decompositions with the 42-comparison adjusted intervals. The two rows for each panel change input and scoring budgets in opposite orders. SciFact\textquotesingle s scoring component is zero by construction because rationale eligibility and the F1 denominators are unchanged at fixed answers.}
\label{tab:path-ci-all}
\begin{center}
\begin{tabular}{lllrr}
\toprule
Dataset & Model & Path order & Input component [CI] & Scoring component [CI] \\
\midrule
FEVER & Qwen & Input first & 1.07 [0.38, 1.86] & 0.10 [-0.20, 0.45] \\
FEVER & Qwen & Scoring first & 1.09 [0.41, 1.87] & 0.08 [0.00, 0.28] \\
FEVER & Llama & Input first & 0.25 [-0.39, 0.93] & 0.09 [-0.13, 0.37] \\
FEVER & Llama & Scoring first & 0.27 [-0.37, 0.96] & 0.07 [0.00, 0.30] \\
FEVEROUS & Qwen & Input first & 0.93 [0.08, 1.87] & 2.91 [0.54, 5.28] \\
FEVEROUS & Qwen & Scoring first & 2.05 [0.85, 3.34] & 1.79 [-0.18, 3.79] \\
FEVEROUS & Llama & Input first & 0.55 [-0.38, 1.50] & 2.56 [0.42, 4.70] \\
FEVEROUS & Llama & Scoring first & 0.88 [-0.10, 1.89] & 2.23 [0.26, 4.21] \\
SciFact & Qwen & Input first & 0.16 [-1.94, 2.06] & 0.00 [0.00, 0.00] \\
SciFact & Qwen & Scoring first & 0.16 [-1.94, 2.06] & 0.00 [0.00, 0.00] \\
SciFact & Llama & Input first & 0.96 [-0.85, 3.49] & 0.00 [0.00, 0.00] \\
SciFact & Llama & Scoring first & 0.96 [-0.85, 3.49] & 0.00 [0.00, 0.00] \\
\bottomrule
\end{tabular}
\end{center}
\end{table}

Table~\ref{tab:path-ci-all} retains all datasets and both orders. In SciFact, 48 of 1,800 source--abstract pairs change their submitted sentence lists between budgets, but none changes complete-rationale eligibility. With answers fixed, the credited-abstract counts and F1 denominators therefore stay the same. Its scoring components and input--scoring interactions are exactly zero for every resample; the zeros describe these submissions, rather than establishing a general absence of context effects.

\section{Reproducibility}
\label{app:reproduce}
The supplement has separate trained-verifier and LLM components. The former reconstructs the retrospective override, checks eight original prediction files, and reproduces all 180 original Study~1 interval rows. An additive post-hoc script computes the five decomposition intervals, gold-label contributions and covariance summaries reported here. These replays use saved answers and scoring flags; original historical weights, model-visible inputs and threshold-selection records remain incomplete.

The LLM component contains raw responses, token/evidence IDs, selected gold, scorer implementations, prompt-building code, family assignments, scoring counts and the original analysis plans. Fully instantiated prompt records and operational execution logs are archived privately; model weights and complete source corpora are excluded. Existing counts and statistical records remain unchanged. The supplement documents each executable check and the boundary between saved-output scoring and reproducing model inference.

\end{document}